\documentclass[11pt,a4paper]{article}
\usepackage[hyperref]{acl2017}
\usepackage{hyperref}
\usepackage{url}
\usepackage{times}
\usepackage{latexsym}
\usepackage{array}
\usepackage{multirow}
\usepackage{amssymb}
\usepackage{amsmath}
\usepackage{mathtools}
\usepackage{amsthm}
\usepackage{xspace}
\usepackage{algorithm}
\usepackage{tikz-dependency}
\usepackage{cancel}
\usepackage{wrapfig}
\usepackage{balance}

\usepackage{graphicx}
\usepackage{caption}
\usepackage{subcaption}

\newcommand{\notes}[1]{}%{\it {\small {#1}}}}

\theoremstyle{definition}

\theoremstyle{plain}

\newcommand{\veca}{\ensuremath{\mathbf{a}}}

\newcommand{\vecz}{\ensuremath{\mathbf{z}}}
\newcommand{\vech}{\ensuremath{\mathbf{h}}}
\newcommand{\vecW}{\ensuremath{\mathbf{W}}}

\newcommand{\ith}[1]{\ensuremath{i^{{th}}}}

\newcount\permx
\newcount\permy
\def\permdot#1#2{
\permx=#1 \advance\permx by-1
\permy=#2 \advance\permy by-1
\psframe[fillcolor=black, fillstyle=solid]
(\permx,\permy)(#1, #2)
}

\newcommand{\boxnum}[1]{{\setlength{\fboxsep}{1pt}\raisebox{1pt}{\hspace{1pt}\fbox{\tiny #1}\hspace{1pt}}}}
\newcommand{\ind}[1]{\ensuremath{_{\kern-0.5pt\boxnum{#1}}}}

\newcommand{\vecx}{\ensuremath{\mathbf{x}}\xspace}
\newcommand{\vecw}{\ensuremath{\mathbf{w}}\xspace}

\newcommand{\vecR}{\ensuremath{\mathbb{R}}\xspace}

\newcommand{\smallnt}[1]{\ensuremath{_{\mbox{\tiny PP}}}\xspace}

\newcommand{\pseudocode}{Algorithm}
\floatname{algorithm}{\pseudocode}

\iffalse

\else

\fi

\newcommand{\Yahoo}{{\sc Yahoo}\xspace}
\newcommand{\TREC}{{\sc TREC}\xspace}
\newcommand{\DMV}{{\sc DMV}\xspace}
\newcommand{\Insurance}{{\sc Insurance}\xspace}

\newcommand{\KL}{\ensuremath{\mathit{KL}}\xspace}

\def\cite{\citep} %%%%%%%%%%%%%%%%%%%%%% natbib
\def\namecite{\cite}

\title{Group Sparse CNNs for Question Classification with Answer Sets}

\author{Mingbo Ma \quad Liang Huang \\
School of EECS\\
Oregon State University\\
Corvallis, OR 97331, USA \\
\tt \small {{\{mam,liang.huang\}@oregonstate.edu}} \\
\And
Bing Xiang \quad Bowen Zhou \\
IBM Watson Group \\
T.~J.~Watson Research Center\\
Yorktown Heights, NY 10598, USA \\
\tt \small {{\{bingxia,zhou\}@us.ibm.com} }\\
}

\aclfinalcopy % Uncomment for camera-ready version

\begin{document}

\maketitle

\begin{abstract}
Question classification is an important task with wide applications.
However, traditional techniques treat questions as general sentences, ignoring the corresponding answer data.
In order to consider answer information into question modeling, 
we first introduce novel group sparse autoencoders 
which refine question representation by utilizing group information in the answer set.
We then propose novel group sparse CNNs 
which naturally learn question representation with respect
to their answers by implanting group sparse autoencoders into traditional CNNs.
The proposed model significantly outperform strong baselines on four datasets.
%\vspace{-0.15cm}
\end{abstract}

%!TEX root = main.tex
%\vspace{-3mm}
\section{Introduction}
\label{sec:intro}

Question classification has applications
in many domains ranging from question answering to dialog systems, 
and has been increasingly popular in recent years.
%question classification has received much attention in recent years.
Several recent efforts \cite{kim:2014,blunsom:2014,ma+:2015} treat questions as general sentences and employ Convolutional Neural Networks (CNNs) to achieve
remarkably strong performance
in the TREC question classification task.
%% as well as other sentence classification
%% tasks such as sentiment analysis.

\iffalse

Most existing approaches to this problem
%however,
simply use existing sentence modeling frameworks
and treat questions as general sentences,
without any special treatment. % tailored to questions.
For example, %% neural network-based sentence modeling frameworks have received  tremendous attention due to the impressive performance and little manual feature engineering.
%In particular,
several recent efforts employ Convolutional Neural Networks (CNNs)
to achieve remarkably strong performance
in the TREC question classification task
as well as other sentence classification
tasks such as sentiment analysis \cite{kim:2014,blunsom:2014,ma+:2015}.

\fi

We argue, however, that
those general sentence modeling frameworks neglect two unique
properties of question classification. 
First, different from the flat and coarse categories in most sentence classification tasks (i.e.~sentimental classification), % or sarcasm detection), 
question classes often have a hierarchical structure such as those from the New York State DMV FAQ\footnote{Crawled from {\scriptsize\url{http://nysdmv.custhelp.com/app/home}}.
This data and our code will be at {\scriptsize\url{http://github.com/cosmmb}}.} (see Fig.~\ref{fig:cate}).
Another unique aspect of question classification is the well prepared answers %with detailed descriptions or instructions 
for each question  or question category.
These answer sets generally cover a larger vocabulary (than the questions themselves) and provide richer information %  carry more distinctive semantic meanings 
for each class.
We believe there is a great potential to enhance question representation with extra information from corresponding answer sets.
% not found in other sentence classification tasks (such as sentimental classification or sarcasm detection), which we detail below:

% \begin{itemize}
% \item 
% The categories for most sentence classification tasks 
% are flat and coarse
% (notable exceptions such as the Reuters Corpus RCV1 \cite{lewis+:2004} notwithstanding), 
% and in many cases, even binary (i.e.~sarcasm detection).
% However, question sentences commonly belong to multiple categories,
% and these categories often have a hierarchical (tree or DAG) structure such as those from the New York State DMV FAQ section \footnote{\scriptsize\url{http://nysdmv.custhelp.com/app/home}} in Fig.~\ref{fig:cate}.

% \item 
% Question sentences from different categories often share similar information or language patterns. This phenomenon becomes more obvious when categories are hierarchical. Fig.~\ref{fig:example} shows one example of questions sharing similar information from different categories. This cross-category shared patterns are not only shown in questions but can also be found in answers corresponding to these questions.

% \item 
% Another unique characteristic for question classification is the well prepared answer set with detailed descriptions or instructions 
% for each corresponding question category.
% These answer sets generally cover a broader range of vocabulary (than the questions themselves) and carry more distinctive semantic meanings for each class.
% We believe there is great potential to enhance the representation of questions with extra information from corresponding answer sets.
% \end{itemize}
\begin{figure}
\begin{center}
\scalebox{0.98}{
\noindent \fbox{\parbox{0.47\textwidth}{%
\textbf{1: Driver License/Permit/Non-Driver ID}\\
a: \textit{Apply for original} \qquad \qquad   \; \; \, (49 questions)\\
b: \textit{Renew or replace} \qquad \qquad \; \; \,  (24 questions)\\[-0.2cm]
...\\
\textbf{2: Vehicle Registrations and Insurance}\\
a: \textit{Buy, sell, or transfer a vehicle} \, \;(22 questions)\\
b: \textit{Reg. and title requirements} \; \; \; \, (42 questions) \\[-0.2cm]
...\\
\textbf{3: Driving Record / Tickets / Points}\\
...
}}
}
\end{center}
%\vspace{-3mm}
\caption{Examples from NYDMV FAQs. There are 8 top-level categories,
47 sub-categories, and 537 questions (among them 388 are {\em unique};
many questions fall into multiple categories).}
\label{fig:cate}
%\vspace{-6mm}
\end{figure} 

To exploit the hierarchical and overlapping structures in question categories
and extra information from answer sets,
we consider dictionary learning %\cite{Aharon05k-svd:design,Roth05fieldsof,Lee07efficientsparse,Candes+:2008,Kreutz:2003,Rubin:2010} 
\cite{Candes+:2008,Rubin:2010}
which is a common approach for representing samples from 
many correlated groups with external information. This learning procedure first builds a dictionary with a series of grouped bases. 
These bases can be initialized randomly or from external data (from the answer set in our case) and optimized during training through Sparse Group Lasso (SGL) \cite{Simon13asparse-group}. 

To apply dictionary learning to CNN,
%Motivated from above, 
we first develop a neural version of SGL, {\em Group Sparse Autoencoders} (GSAs),
which to the best of our knowledge, is the first full neural model with group sparse constraints.
 % which is a neural-based version of SGL. 
% The objective of GSA and SGL are very similar. 
The encoding matrix of GSA (like the dictionary in SGL) is grouped into different categories. 
The bases in different groups can be either initialized randomly or by the sentences in corresponding answer categories. 
Each question sentence will be reconstructed by a few bases within a few groups. 
GSA can use either linear or nonlinear encoding or decoding while SGL is restricted to be linear.
Eventually, %Based on GSA, 
to model questions with sparsity,
%In order to incorporate both advantages from GSA and CNNs, 
we further propose novel {\em Group Sparse Convolutional Neural Networks} (GSCNNs) by implanting the GSA onto CNNs, 
essentially enforcing group sparsity between the convolutional and classification layers. 
%% GSCNNs are jointly trained %end-to-end 
%% neural-based framework for getting question representations with group sparse constraint from both answer and question sets. 
This framework is a jointly trained %end-to-end 
neural model to learn question representation with group sparse constraints from both question and answer sets. 

% Experiments show significant improvements over strong baselines
% on four datasets.

% \begin{figure}
% \begin{center}
% \noindent \fbox{\parbox{0.45\textwidth}{%
% \textbf{Category: Finance}\\
% Q: How to get a personal loan from the bank?\\
% \textbf{Category: Education}\\
% Q: What are the steps for applying for student loan?}}
% \end{center}
% \caption{Examples of questions from two different categories. These questions ask for the similar problem even if they are in different classes. Their answers also contain similar information.}
% \label{fig:example}
% \vspace{-3mm}
% \end{figure}

%!TEX root = main.tex

\section{Group Sparse Autoencoders}
\label{sec:GSA}

\subsection{Sparse Autoencoders}
Autoencoder \cite{NIPS2006_3048} is an unsupervised neural network which learns the hidden representations from data. %of input samples.
When the number of hidden units is large %and redundant 
(e.g., bigger than input dimension),
we can still discover the underlying structure by imposing sparsity constraints, % on the network.
%To achieve this, 
using sparse autoencoders (SAE) \cite{andrew_sparse}:% shows interesting results of getting visualization of the hidden layers.
%The objective function of SAE is: %defined as follows:
%\footnote{\small{for more details please refer to \href{http://web.stanford.edu/class/cs294a/sparseAutoencoder.pdf}{the paper} \cite{andrew_sparse}}} %\url{http://web.stanford.edu/class/cs294a/sparseAutoencoder.pdf}}}
%\vspace{-6mm}
\begin{equation}\label{eq:loss_sparse}
%\small
    J_\text{sparse}(\rho) = J + \alpha \sum_{j=1}^s \KL(\rho \| \hat{\rho}_j)
\end{equation}
%\vspace{-3mm}
where $J$ is the autoencoder reconstruction loss, 
$\rho$ is the desired sparsity level which is small,
and thus $J_\text{sparse}(\rho)$ is the sparsity-constrained version of loss $J$.
Here $\alpha$ is the weight of the sparsity penalty term defined below:
%\vspace{-6mm}
\begin{equation}\label{eq:KL_sparse}
%\small
    \KL(\rho \| \hat{\rho}_j)=  \rho\log\frac{\rho}{\hat{\rho}_j} + (1-\rho)\log\frac{1-\rho}{1-\hat{\rho}_j}
\end{equation}
where \[\hat{\rho}_j = \frac{1}{m}\sum_{i=1}^m h_j^i\] represents the average activation of hidden unit $j$
over $m$ examples  (SAE assumes the input features are correlated). % value.

As described above, SAE has a similar objective to traditional sparse coding which tries to find sparse representations for input samples. 
Besides applying simple sparse constraints to the network,
group sparse constraints is also desired when the class categories are structured and overlapped.  
Inspired by group sparse lasso \cite{Yuan06modelselection} and sparse group lasso \cite{Simon13asparse-group}, we propose a novel architecture below.
%, Group Sparse Autoencoders (GSA) in this paper. 

\begin{figure*}[h]
    \centering
    \begin{subfigure}[b]{0.2\textwidth}
      \raisebox{0.5cm}{\includegraphics[width=\textwidth]{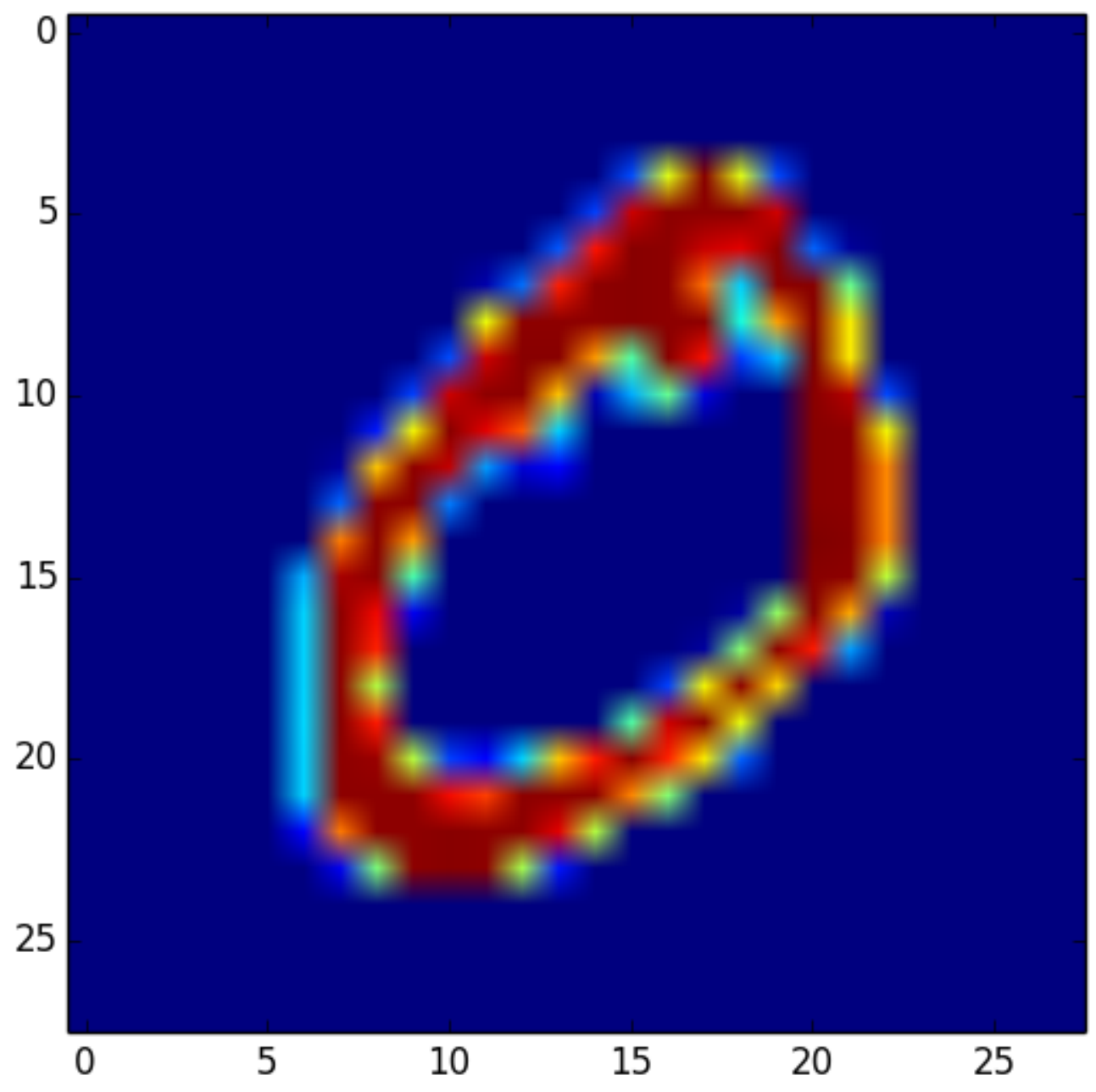}}
        \caption{}
        \label{fig:img}
    \end{subfigure}
    ~ %add desired spacing between images, e. g. ~, \quad, \qquad, \hfill etc. 
    \begin{subfigure}[b]{0.45\textwidth}
        \includegraphics[width=\textwidth,height=4.5cm]{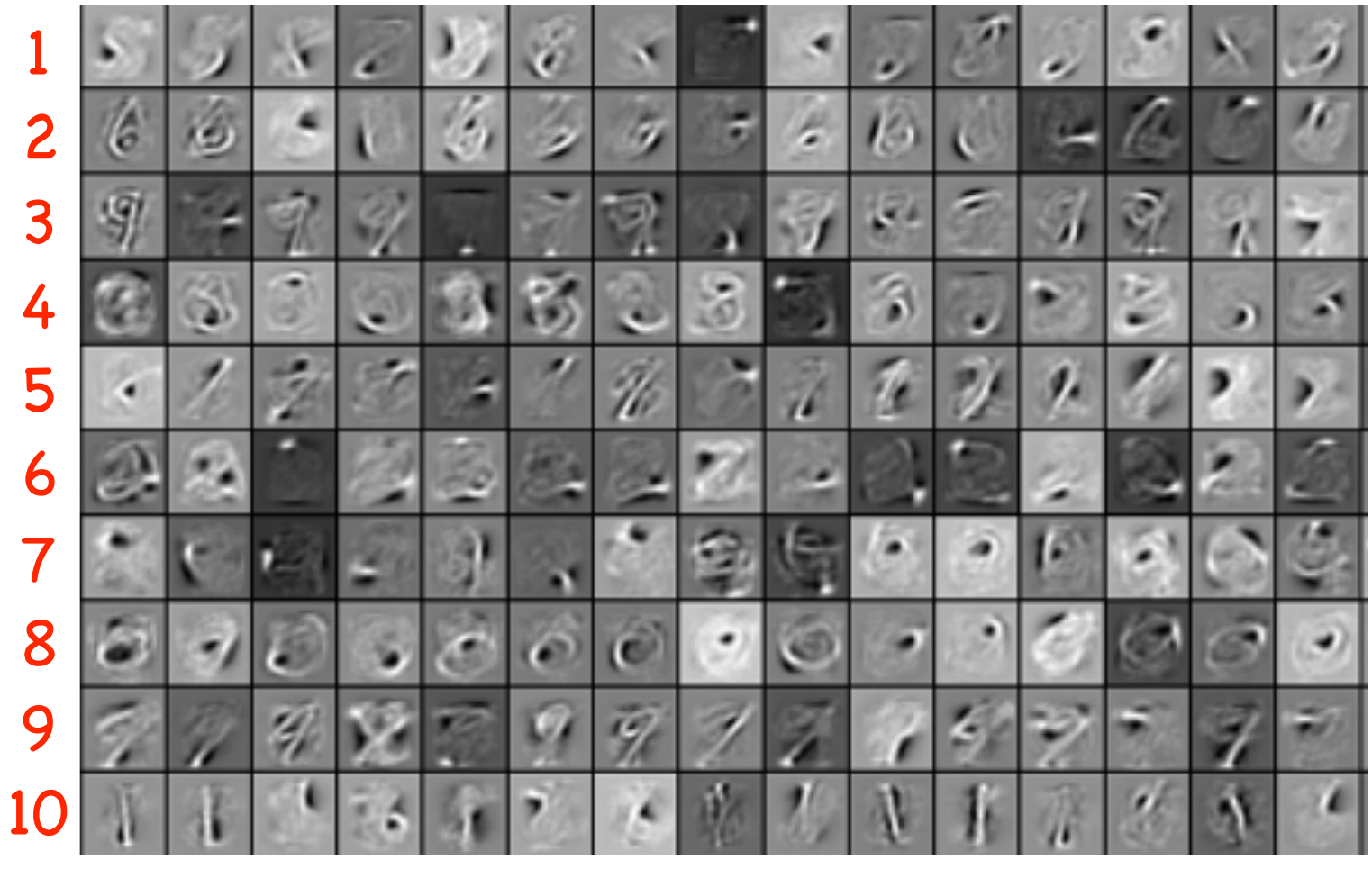}
        \caption{}
        \label{fig:w1}
    \end{subfigure}
    ~ %add desired spacing between images, e. g. ~, \quad, \qquad, \hfill etc. 
    \begin{subfigure}[b]{0.28\textwidth}
        \raisebox{0.05cm}{\includegraphics[width=\textwidth,height=4.4cm]{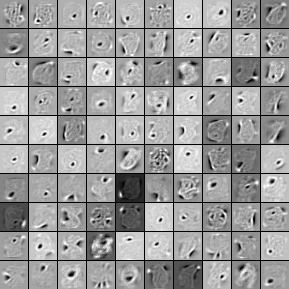}}
        \caption{}
        \label{fig:w2}
    \end{subfigure}
    \caption{The input figure with hand written digit $0$ is shown in (a). Figure (b) is the visualization of trained projection matrix $\vecW$ on MNIST dataset. Different rows represent different groups of $\vecW$ in Eq.~\ref{eq:loss_sgl}. 
For each group, we only show the first 15 (out of 50) bases. The red numbers on the left side are the indices of 10 different groups. Figure (c) is the projection matrix from basic autoencoders.% The differences are easier to see in pdf.
}\label{fig:vis}
\end{figure*}

\subsection{Group Sparse Autoencoders}
Group Sparse Autoencoder (GSA), unlike SAE, categorizes the weight matrix into different groups. 
For a given input, GSA reconstructs the input signal with the activations from only a few groups. 
Similar to the average activation $\hat{\rho}_j$ for sparse autoencoders, GSA defines each grouped average activation for the hidden layer as follows: 
\begin{equation}\label{eq:eta_hat_j}
    \hat{\eta}_p = \frac{1}{mg}\sum_{i=1}^m \sum_{l=1}^g  \| h^{i}_{p,l} \| _2
\end{equation}
where $g$ represents the size of each group, and $\hat{\eta}_j$ first sums up all the activations within $p^{th}$ group, then computes the average $p^{th}$ group respond across different samples' hidden activations. 

Similar to Eq.~\ref{eq:KL_sparse}, we also use \KL divergence to measure the difference between estimated intra-group activation and global group sparsity: % as follows:
\begin{equation}\label{eq:KL_grop}
    \KL(\eta \| \hat{\eta}_p) = \eta\log\frac{\eta}{\hat{\eta}_p} + (1-\eta)\log\frac{1-\eta}{1-\hat{\eta}_p}
\end{equation}
where $G$ is the number of groups. Then the objective function of GSA is: % defined as follows:
\begin{equation}\label{eq:loss_sgl}
\begin{split}
J_\text{groupsparse}(\rho, \eta) = J & + \alpha \sum_{j=1}^s \KL(\rho \| \hat{\rho}_j) \\
& + \beta \sum_{p=1}^G \KL(\eta \| \hat{\eta}_p)
\end{split}
\end{equation}
where $\rho$ and $\eta$ are constant scalars which are our target sparsity and group-sparsity levels, resp.
When $\alpha$ is set to zero, GSA only considers the structure between difference groups. When $\beta$ is set to zero, GSA is reduced to SAE.  
%In some cases, if both inter- and intra-group sparsities are preferred we need to set both $\alpha$ and $\beta$ to positive values. 

\subsection{Visualizing Group Sparse Autoencoders}

% \begin{figure*}
%     \centering
%     \begin{subfigure}[b]{0.4\textwidth}
%         \includegraphics[width=\textwidth]{figs/2.pdf}
%         \caption{}
%         \label{fig:gull}
%     \end{subfigure}
%     ~ %add desired spacing between images, e. g. ~, \quad, \qquad, \hfill etc. 
%       %(or a blank line to force the subfigure onto a new line)
%     \begin{subfigure}[b]{0.32\textwidth}
%         \includegraphics[width=\textwidth,height=3.9cm]{figs/act.pdf}
%         \caption{}
%         \label{fig:tiger}
%     \end{subfigure}
%     ~ %add desired spacing between images, e. g. ~, \quad, \qquad, \hfill etc. 
%     %(or a blank line to force the subfigure onto a new line)
%     \begin{subfigure}[b]{0.24\textwidth}
%         \includegraphics[width=\textwidth]{figs/3.png}
%         \caption{}
%         \label{fig:mouse} 
%     \end{subfigure}
% %\vspace{-0.2cm}
%     \caption{(a): the visualization of trained projection matrix $\vecW$ on MNIST dataset. Different rows represent different groups of $\vecW$ in Eq.~\ref{eq:loss_sgl}. 
% For each group, we only show the first 15 (out of 50) bases. The red numbers on the left side are the indices of 10 different groups. % (10 in total). 
% (b): the hidden activations $\vech$ with respect to the input image (the red numbers correspond to the indices in (a)). 
% (c): the projection matrix from basic autoencoders. The differences are easier to see in pdf.}\label{fig:vis}
% %\vspace{-.3cm}
% \end{figure*}

\newcommand{\myspace}{\hspace{0.175cm}$\mid$\hspace{0.175cm}}
\begin{figure*}
%\centering
%\hspace{-0.3cm}
%\begin{tabular}{p{.4cm}p{0.4\textwidth}}
\begin{tabular}{cc}
%\raisebox{2cm}{(a)} &
  \begin{subfigure}[b]{0.5\textwidth}
     %\centering
     \includegraphics[width=.85\textwidth,height=4.2cm]{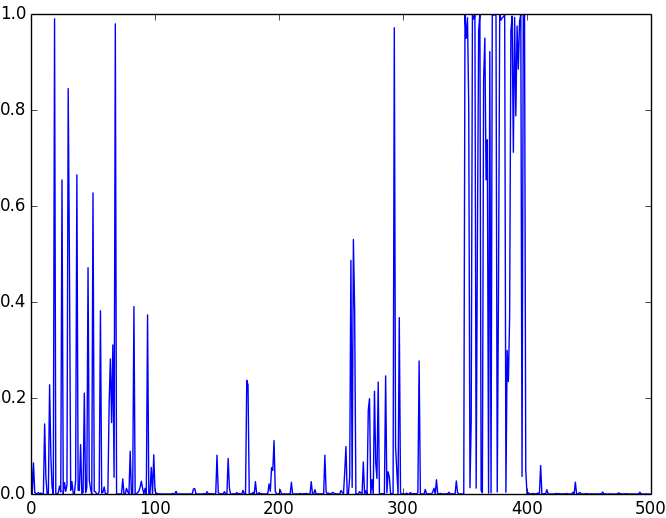}\\[-0.1cm]
     {\color{red}\hspace{.45cm}1\myspace 2\myspace 3\myspace 4\myspace 5\myspace 6\myspace 7\myspace 8\myspace 9\myspace 10}
     %\caption{}
     %\label{fig:Ng1} 
   \end{subfigure}
&
\begin{subfigure}[b]{0.5\textwidth}
     \raisebox{.3cm}{\includegraphics[width=.85\textwidth,height=4.2cm]{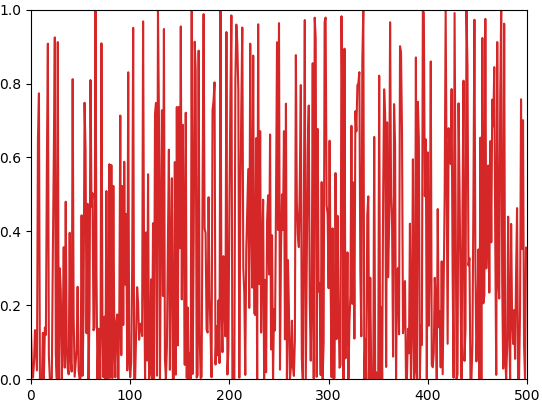}}
     %\caption{}
     %\label{fig:Ng2}
\end{subfigure}
\\%[0.5cm]
\hspace{-1cm}(a) & \hspace{-1cm}(b)
%\raisebox{2cm}{(b)} &   
\end{tabular}
\caption{(a): the hidden activations $\vech$ for the input image in Fig.~\ref{fig:vis}(a). 
The red numbers corresponds to the index in Fig.~\ref{fig:vis}(b). 
(b): the hidden activations $\vech$ for the same input image from basic autoencoders.}\label{fig:act}
\end{figure*}

% \begin{figure}[!htbp]
% \centering
% \includegraphics[width=0.45\textwidth,height=5cm]{figs/act.pdf}
% \caption{The hidden activations $\vech$ with respect to the input image in Fig.~\ref{fig:vis}(a). the red numbers corresponds to the index in Fig.~\ref{fig:vis}(b). These activations come from $10$ different groups. The group size here is $50$.}
% \label{fig:act}
% \end{figure}

In order to have a better understanding of GSA, %of how the GSA behaves, 
we use the MNIST dataset to visualize GSA's internal parameters. % of GSA.
% MNIST dataset is used here for this visualization experiments. 
Fig.~\ref{fig:vis} and Fig.~\ref{fig:act} illustrate the projection matrix and the corresponding hidden activations.
We use 10,000 training samples. We set the size of the hidden layer to 500 with 10 groups. Fig.~\ref{fig:vis}(a) visualizes the input image for hand written digit $0$.% for GSA. 
% We set the intra-group sparsity $\rho$ equal to $0.3$ and inter-group sparsity $\eta$ equal to $0.2$. $\alpha$ and $\beta$ are equal to 1. 
% On the other hand, we also train the same $10,000$ examples on basic autoencoders with random noise added to the input signal (denoising autoencoders \cite{VincentPLarochelleH2008}) for better hidden information extraction. 
% We add the same $30\%$ random noise into both models. 
% Note that the group size of this experiments does not have to be set to $10$. Since this is the image dataset with digit numbers, we may use fewer groups to train GSA.

In Fig.~\ref{fig:vis}(b), we find similar patterns within each group. For example, group 8 has different forms of digit 0, 
and group 9 includes different forms of digit 7. However, it is difficult to see any meaningful patterns from the projection matrix of basic autoencoders in Fig.~\ref{fig:vis}(c). 

Fig.~\ref{fig:act}(a) shows the hidden activations with respect to the input image of digit 0. 
The patterns of the 10$^{th}$ row in Fig.~\ref{fig:vis}(b) are very similar to digit $1$ which is very different from digit $0$ in shape. Therefore, there is no activation in group 10 in Fig.~\ref{fig:act}(a).
The majority of hidden layer activations are in groups 1, 2, 6 and 8,
with group 8 being the most significant. % activations. 
When compared to the projection matrix visualization in Fig.~\ref{fig:vis}(b), these results are reasonable since the 8$^{th}$ row has the most similar patterns of digit 0.
However, we could not find any meaningful pattern from the hidden activations of basic autoencoder as shown in Fig.~\ref{fig:act}(b).

GSA could be directly applied to small image data (e.g.~MINIST dataset) for pre-training. 
However, in tasks which prefer dense semantic representations (e.g.~sentence classification), we still need CNNs to learn the sentence representation automatically. 
%In this scenario, 
In order to combine advantages from GSA and CNNs, we propose Group Sparse Convolutional Neural Networks below.%in the following section.

%!TEX root = main.tex
%\vspace{-0.2cm}
\section{Group Sparse CNNs}
%\vspace{-0.2cm}
\label{sec:GSCNN}

%\begin{wrapfigure}[19]{l}[\dimexpr\columnwidth+\columnsep\relax]{12cm}
%\begin{wrapfigure}{l}{0.7\textwidth}%[ht!]
\begin{figure*}
\centering
\includegraphics[width=0.9\textwidth]{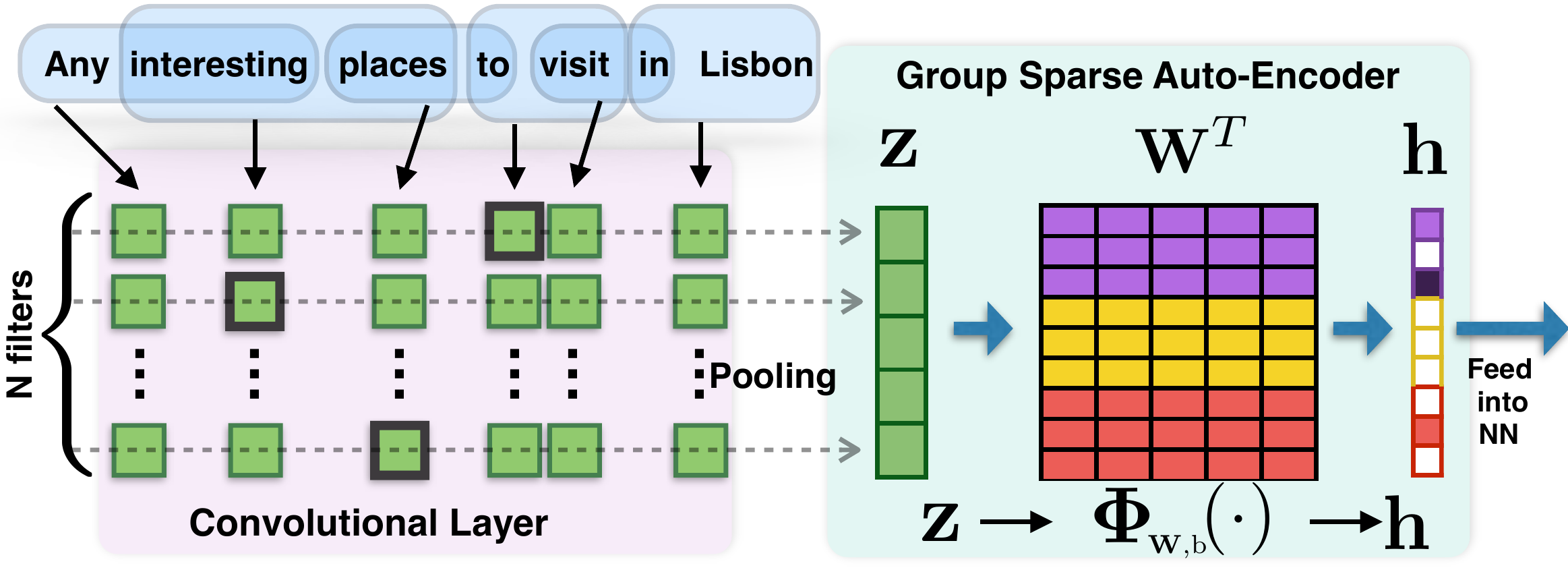}
\caption{Group Sparse CNN. We add an extra dictionary learning layer between sentence representation $\vecz$ and the final classification layer. 
%Sentence representation after convolutional layer is denoted as $\vecz$, and 
$\vecW$ is the projection matrix (functions as a dictionary) 
that converts $\vecz$ to the group sparse representation $\vech$ (Eq.~\ref{eq:loss_sgl}).
%Hidden group sparse representation for question sentence is denoted as $\vech$. 
Different colors in the projection matrix represent different groups. 
We show $\vecW^\intercal$ instead of $\vecW$ for presentation purposes. 
Darker colors in $\vech$ mean larger values and white means zero.}
\label{fig:model}
%\vspace{-0.1cm}
\end{figure*}
%\end{wrapfigure}

% CNNs were first proposed by \cite{LeCun95comparisonof} in computer vision. 
% For a given image, CNNs apply convolution kernels on a series of continuous areas on images. 
% This concept was first adapted to NLP by \cite{collobert+:2011}. 
% Recently, many CNNs-based techniques 
% achieve great successes in sentence modeling and classification
% \cite{kim:2014,blunsom:2014,ma+:2015}. 
% For simplicity, we use the sequential CNNs \citep{kim:2014} as our baseline.

CNNs were first proposed by \namecite{LeCun95comparisonof} in computer vision and
adapted to NLP by \cite{collobert+:2011}.
Recently, many CNN-based techniques have
achieved great successes in sentence modeling and classification
\cite{kim:2014,blunsom:2014}.
% For simplicity, we use the sequential CNNs \citep{kim:2014} as our baseline.

Following sequential CNNs, one dimensional convolutions operate the convolution kernel in sequential order $\vecx_{i,j} =\vecx_i \oplus   \vecx_{i+1}\oplus \cdots \oplus  \vecx_{i+j}$, where~$\vecx_i \in \vecR^e$ represents the $e$ dimensional word representation for the $i$-th word in the sentence, and~$\oplus$ is the concatenation operator. Therefore $\vecx_{i,j}$ refers to concatenated word vector from the $i$-th word to the $(i+j)$-th word in sentence.

A convolution operates a filter $\mathbf{w} \in  \vecR^{n \times e}$ to a window of $n$ words $\vecx_{i,i+n}$ with bias term $b'$ by $a_i = \sigma (\vecw \cdot  \vecx_{i,i+n}+b' )$ 
with non-linear activation function $\sigma$ to produce a new feature.
The filter $\vecw$ is applied to each word in the sentence, generating the feature map $\veca= [a_1, a_2, \cdots, a_L]$ where $L$ is the sentence length.
We then use $\hat{a} = \max \{ \veca \}$ to represent the entire feature map after max-pooling.

% The convolution described in Eq.~\ref{eq:con_def} can be regarded as feature detection: more similar patterns will return higher activation. In sequential CNNs, max-over-time pooling \cite{collobert+:2011,kim:2014} operates over the feature map to get the maximum activation $\hat{a} = \max \{ \veca \}$ representing the entire feature map. The idea is to detect the strongest activation over time. This pooling strategy also naturally deals with sentence length variations.

In order to capture different aspects of patterns, CNNs usually randomly initialize a set of filters with different sizes and values. 
Each filter will generate a feature as described above. 
To take all the features generated by $N$ different filters into count, we use $\vecz = [\hat{a_1}, \cdots, \hat{a_N}]$ as the final representation.
In conventional CNNs, this $\vecz$ will be directly fed into classifiers after the sentence representation is obtained, e.g.~fully connected neural networks \cite{kim:2014}. There is no easy way for CNNs to explore the possible hidden representations with underlaying structures.

In order to exploit these structures, % obtain the hidden representations for each sentence representation, 
we propose Group Sparse Convolutional Neural Networks (GSCNNs) by placing one extra layer between the convolutional and the classification layers. 
This extra layer mimics the functionality of GSA from Section~\ref{sec:GSA}.
Shown in Fig.~\ref{fig:model},
%The convolutional layer follows the traditional convolutional process. % described above.
after the conventional convolutional layer, we get the feature map $\vecz$ for each sentence. 
%The feature maps $\vecz$ is treated as the feature representation for each sentence. 
In stead of directly feeding it into a fully connected neural network for classification, 
we enforce the group sparse constraint on $\vecz$ in a way similar to the group sparse constraints on hidden layer in GSA from Sec.~\ref{sec:GSA}. %like the group sparse constraint we have on $h$ in Eq.~\ref{eq:loss_sgl}. 
Then, we use the sparse hidden representation $\vech$ in Eq.~\ref{eq:loss_sgl} as the new sentence representation,
which is then fed %. The last step is feeding the hidden representation $\vech$ 
into a fully connected neural network for classification. The parameters $\vecW$ in Eq.~\ref{eq:loss_sgl} will also be fine tunned during the last step.

% In order to improve the robustness of the hidden representation and prevent it from simply learning the identity, we follow the idea of decisioning autoencoders \cite{VincentPLarochelleH2008} to add random noise (10\% in our experiments) into $\vecz$. The training process of our model is similar to the training process in stack autoencoders \cite{NIPS2006_3048}. 

% In order to prevent the co-adaptation of the hidden unites, we employ random dropout on penultimate layer \cite{hinton:2014}. We set the drop out rate as $0.5$ and learning rate as $0.95$ by default. 
% In our experiments, training is done through stochastic gradient descent over shuffled mini-batches with the Adadelta update rule \cite{zeiler:2012}. All the settings of the CNNs are the same as the settings in \cite{kim:2014}.
Different ways of initializing the projection matrix in Eq.~\ref{eq:loss_sgl} can be summarized below: % as the followings:

\begin{itemize}

\item \textbf{Random Initialization}: When there is no answer corpus available, we first randomly initialize $N$ vectors 
%(usually $N \gg s$) 
to represent the group information from the answer set. 
Then we cluster these $N$ vectors into $G$ categories with $g$ centroids for each category. 
These centroids from different categories will be the initialized bases for projection matrix $\vecW$ which will be learned during training.
%\vspace{-0.1cm}
\item \textbf{Initialization from Questions}: Instead of using random initialized vectors, we can also use question sentences for initializing 
the projection matrix when the answer set is not available. We need to pre-train the sentences with CNNs to get the sentence representation. 
We then select $G$ largest categories in terms of number of question sentences. Then we get $g$ centroids from each category by $k$-means. 
We concatenate these $G\times g$ vectors to form the projection matrix. %We need to pre-train the sentence with CNNs to get the sentence representation.  
%\vspace{-0.1cm}
\item \textbf{Initialization from Answers}: This is the most ideal case. We follow the same procedure as above, with the only difference being using the answer sentences in place of question sentences to pre-train the CNNs. % to get answer sentence representation.
%\vspace{-0.1cm}
\end{itemize}

%!TEX root = main.tex
%\vspace{-0.2cm}
\section{Experiments}
%\vspace{-0.3cm}
% \subsection{Performance on CQA}
\label{sec:exp1}

Since there is little effort to use answer sets in question classification, we did not find any suitable datasets which are publicly available. 
We collected two datasets ourselves and also used two other well-known ones. %datasets in our experiments.   
These datasets are summarized in Table~\ref{tb:data}. 
\Insurance is a private dataset we collected from a car insurance company's website.
Each question is classified into  319 classes with corresponding answer data.
All questions which belong to the same category share the same answers. 
The \DMV dataset is collected from New York State the \DMV's FAQ website. %We will make this data publicly available in the future.
The \Yahoo~Ans dataset is only a subset of the original publicly available \Yahoo~Answers dataset \cite{Fleming,Shah:2010:EPA:1835449.1835518}. % which is \href{http://webscope.sandbox.yahoo.com/catalog.php?datatype=l}{publicly available}.
Though not very suitable for our framework, 
we still included the frequently used \TREC dataset (factoid question type classification) for comparison.

We only compare our model's performance with CNNs for two following reasons: we consider our ``group sparsity''
as a modification to the general CNNs for grouped feature selection. 
This idea is orthogonal to any other CNN-based models and 
can be easily applied to them; 
in addition, as discussed in Sec.~\ref{sec:intro}, 
we did not find any other model in comparison with solving question classification tasks with answer sets.

%% The datasets we use in the experiments require the label information for both questions and answers. 
%% Besides that, similar with websites' FAQ section, all the questions which belong to the same category share the same answer sets.
%% Among the above the four datasets, only the \Insurance and \DMV datasets are suitable for our model.
%% The questions which fall into the same category have different answers in \Yahoo dataset. 

There is crucial difference between the \Insurance and \DMV datasets on one hand and the \Yahoo set on the other.
In \Insurance and \DMV, all questions in the same (sub)category share the same answers,
whereas \Yahoo provides individual answers to each question.

\begin{table}[t]
\centering
\scalebox{0.75}{
\begin{tabular}{ |l|l|l|l|l|l|c| }
\hline
            Datasets& $C_t$ &  $C_s$ &  $N_{data}$   & $N_{test}$ & $N_{ans}$ & Multi-label\\ 
\hline
                \TREC    &  6  &    $50$   & 5952   &  500   & - & No\\
\hline
               \Insurance  & - & 319      &  1580  &    303    & 2176 & Yes  \\
\hline
                \DMV     &  8   &  47 & 388   &  50  & 2859 & Yes\\
\hline
                 \Yahoo Ans      & 27  &  678       &  8871  &  3027  & 10365  & No \\
\hline
\end{tabular}
}
%\vspace{-0.1cm}
\caption{Summary of datasets. $C_t$ and $C_s$ are the numbers of top-level and sub- categories, resp.
%and $C_s$ is the number of sub-categories. 
%Note we only do top level classification on \TREC. 
$N_{\mathrm{data}}$, $N_{\mathrm{test}}$, $N_{\mathrm{ans}}$ are the sizes of data set, test set, and answer set, resp.
Multilabel means each question can belong to multiple categories.}
\label{tb:data}
%\vspace{-0.1cm}
\end{table}

\begin{table}%[!htbp]
\centering
\scalebox{0.8}{
\begin{tabular}{|c|rrr|r r|r|}
\hline
\multirow{2}{*}{} & \multirow{2}{*}{\!\!\TREC\!\!} &\multirow{2}{*}{\!\!\!\sc Insur.\!\!} & \multirow{2}{*}{\!\!\!\DMV\!\!}
& \multicolumn{3}{c|}{\Yahoo dataset}\\
\cline{5-7}
 & & & & sub & top & unseen \\
%\cline{5-7}
\hline
CNN$^\dagger$  & 93.6 & 51.2& 60& 20.8& 53.9&47\\
%\hline
+sparsity$^\ddagger$   & 93.2 & 51.4& 62& 20.2& 54.2&46\\
\hline
\hline
$\vecW_R$ & 93.8& 53.5& 62& 21.8& 54.5&48\\
%\hline
$\vecW_Q$ & $\textbf{94.2}$& 53.8& 64& 22.1& 54.1&48\\
%\hline
$\vecW_A$ & - & $\textbf{55.4}$& $\textbf{66}$ & $\textbf{22.2}$ & $\textbf{55.8}$ & $\textbf{53}$\\
\hline
\end{tabular}
}
%\vspace{-0.2cm}
\caption{Experimental results. Baselines: $^\dagger$sequential CNNs ($\alpha=\beta=0$ in Eq.~\ref{eq:loss_sgl}), $^\ddagger$CNNs with global sparsity ($\beta=0$). 
$\vecW_R$: randomly initialized projection matrix. $\vecW_Q$: question-initialized projection matrix. 
$\vecW_A$: answer set-initialized projection matrix. % represents the performance of the model whose projection matrix is initialized by answer set. 
There are three different classification settings for \Yahoo: subcategory, top-level category, and top-level accuracies on unseen sub-labels.}
\label{tb:results}
\vspace{-0.2cm}
\end{table}

%Note that projection matrix will be updated during training for better classification performance.

%In the cases of single-label classification tasks (\TREC and \Yahoo), we set the last layer as softmax-layer which tries to get one unique peaky choice across all other labels. 
For multi-label classification (\Insurance and \DMV), we replace the softmax layer in CNNs with a sigmoid layer which predicts each category independently 
while softmax is not. %function has an exclusive property which allows cross influence between categories. 

All experimental results are summarized in Table~\ref{tb:results}. 
The improvements are substantial for \Insurance and \DMV,
but not as significant for \Yahoo and \TREC.
%The improvement on \Yahoo is not as significant as \Insurance and \DMV. 
One reason for this is the questions in \Yahoo/\TREC are shorter, %, sometimes only with 2--3 words. 
%When the sentences become shorter, 
which makes the group information harder to encode. 
Another reason is that each question in \Yahoo/\TREC has a single label, and thus can not fully benefit from group sparse properties.
%\Yahoo-top shows the results of top-level category classification. We map the subcategories back to the top categories and get the results in Table~\ref{tb:results}.

Besides the conventional classification tasks, we also test our proposed model on an unseen-label case. In these experiments, there are a few sub-category labels that are not included in the training data. However, we still hope that our model could still return the correct parent category for these unseen subcategories at test time.
% label into correct parent category based on the model's sub-category estimation. 
In the testing set of \Yahoo dataset, we randomly add 100 questions whose subcategory labels are unseen in training set. The classification results of \Yahoo-unseen in Table~\ref{tb:results} are obtained by mapping the predicted subcategories back to top-level categories. %and check whether the true label's top category match with predicted label's parent category. 
The improvements are substantial due to the group information encoding.

%\input{related}

%\vspace{-.1cm}
\section{Conclusions}% and Future Work}

%In this paper, 
In order to better represent question sentences with answer sets and group structure,
we first presented a novel GSA framework, a neural version of dictionary learning. % and sparse coding models with inter- and intra- group sparse constraints. 
%We also demonstrated GSA's learning ability by visualizing the projection matrix and activations. 
We then proposed group sparse convolutional neural networks by embedding GSA into CNNs,
which result in significantly better question classification over strong baselines. 
%We show that CNNs can benefit from GSA by learning more meaningful representation from dictionary.

\vspace{-.1cm}

\section*{Acknowledgment}
We thank the anonymous reviewers for their suggestions. 
This work is supported in part by 
NSF IIS-1656051,
DARPA FA8750-13-2-0041 (DEFT),
DARPA XAI,
a Google Faculty Research Award,
and an HP Gift.

\balance
\bibliography{acl2017}

\begin{thebibliography}{}
\expandafter\ifx\csname natexlab\endcsname\relax\def\natexlab#1{#1}\fi

\bibitem[{Bengio et~al.(2007)Bengio, Lamblin, Popovici, and
  Larochelle}]{NIPS2006_3048}
Yoshua Bengio, Pascal Lamblin, Dan Popovici, and Hugo Larochelle. 2007.
\newblock Greedy layer-wise training of deep networks.
\newblock In {\em Advances in Neural Information Processing Systems 19\/}.

\bibitem[{Cand\`{e}s and Wakin(2008)}]{Candes+:2008}
Emmanuel~J. Cand\`{e}s and Michael~B. Wakin. 2008.
\newblock \href{http://dx.doi.org/10.1109/msp.2007.914731}{{An Introduction To
  Compressive Sampling}}.
\newblock In {\em Signal Processing Magazine, IEEE\/}. volume~25.
\newblock
  \href{http://dx.doi.org/10.1109/msp.2007.914731}{http://dx.doi.org/10.1109/msp.2007.914731}.

\bibitem[{Collobert et~al.(2011)Collobert, Weston, Bottou, Karlen, Kavukcuoglu,
  and Kuksa}]{collobert+:2011}
R.~Collobert, J.~Weston, L.~Bottou, M.~Karlen, K.~Kavukcuoglu, and P.~Kuksa.
  2011.
\newblock Natural language processing (almost) from scratch.
\newblock In {\em Journal of Machine Learning Research\/}. volume~12, pages
  2493--2537.

\bibitem[{Fleming et~al.(2012)Fleming, Chalmers, and Wakeman}]{Fleming}
Simon Fleming, Dan Chalmers, and Ian Wakeman. 2012.
\newblock A deniable and efficient question and answer service over ad hoc
  social networks.
\newblock In {\em Information Retrieval\/}.

\bibitem[{Kalchbrenner et~al.(2014)Kalchbrenner, Grefenstette, and
  Blunsom}]{blunsom:2014}
Nal Kalchbrenner, Edward Grefenstette, and Phil Blunsom. 2014.
\newblock A convolutional neural network for modelling sentences.
\newblock In {\em Proceedings of the 52nd Annual Meeting of the Association for
  Computational Linguistics\/}. Association for Computational Linguistics.

\bibitem[{Kim(2014)}]{kim:2014}
Yoon Kim. 2014.
\newblock \href{http://www.aclweb.org/anthology/D14-1181}{Convolutional neural
  networks for sentence classification}.
\newblock In {\em Proceedings of the 2014 Conference on Empirical Methods in
  Natural Language Processing (EMNLP)\/}. Association for Computational
  Linguistics, Doha, Qatar, pages 1746--1751.
\newblock
  \href{http://www.aclweb.org/anthology/D14-1181}{http://www.aclweb.org/anthology/D14-1181}.

\bibitem[{LeCun et~al.(1995)LeCun, Jackel, Bottou, Brunot, Cortes, Denker,
  Drucker, Guyon, Müller, Säckinger, Simard, and
  Vapnik}]{LeCun95comparisonof}
Y.~LeCun, L.~Jackel, L.~Bottou, A.~Brunot, C.~Cortes, J.~Denker, H.~Drucker,
  I.~Guyon, U.~Müller, E.~Säckinger, P.~Simard, and V.~Vapnik. 1995.
\newblock Comparison of learning algorithms for handwritten digit recognition.
\newblock In {\em International Conference on Artificial Neural Networks\/}.
  pages 53--60.

\bibitem[{Ma et~al.(2015)Ma, Huang, Xiang, and Zhou}]{ma+:2015}
Mingbo Ma, Liang Huang, Bing Xiang, and Bowen Zhou. 2015.
\newblock Dependency-based convolutional neural networks for sentence
  embedding.
\newblock In {\em Proceedings of ACL 2015\/}.

\bibitem[{Ng(2011)}]{andrew_sparse}
Andrew Ng. 2011.
\newblock Sparse autoencoder.
\newblock In {\em CS294A Lecture notes\/}. Stanford University, page~72.

\bibitem[{Rubinstein et~al.(2010)Rubinstein, Bruckstein, and Elad}]{Rubin:2010}
R.~Rubinstein, A.~M. Bruckstein, and M.~Elad. 2010.
\newblock Dictionaries for sparse representation modeling.
\newblock In {\em Neural Computation\/}.

\bibitem[{Shah and Pomerantz(2010)}]{Shah:2010:EPA:1835449.1835518}
Chirag Shah and Jefferey Pomerantz. 2010.
\newblock Evaluating and predicting answer quality in community qa.
\newblock In {\em Proceedings of the 33rd International ACM SIGIR Conference on
  Research and Development in Information Retrieval\/}. ACM, New York, NY, USA.

\bibitem[{Simon et~al.(2013)Simon, Friedman, Hastie, and
  Tibshirani}]{Simon13asparse-group}
Noah Simon, Jerome Friedman, Trevor Hastie, and Rob Tibshirani. 2013.
\newblock A sparse-group lasso.
\newblock In {\em Journal of Computational and Graphical Statistics\/}.

\bibitem[{Yuan and Lin(2006)}]{Yuan06modelselection}
Ming Yuan and Yi~Lin. 2006.
\newblock Model selection and estimation in regression with grouped variables.
\newblock In {\em Journal of the Royal Statistical Society\/}. volume~68, pages
  49--67.

\end{thebibliography}
\bibliographystyle{acl_natbib}

\end{document}